\documentclass[11pt]{article}
\usepackage[utf8]{inputenc}
\usepackage{amsmath, amssymb, amsthm}
\usepackage{algorithm}
\usepackage{algpseudocode}
\usepackage{geometry}
\usepackage{graphicx}
\usepackage{booktabs}
\usepackage{enumitem}
\usepackage{xcolor}
\usepackage{subcaption}
\usepackage{noto}
\usepackage{hyperref}

\title{Bayesian Inference of Training Dataset Membership}
\author{Yongchao Huang\footnote{Author email: \texttt{yongchao.huang@abdn.ac.uk}}}

\date{May 2025}

\begin{document}

\maketitle

\begin{abstract}
Determining whether a dataset was part of a machine learning model's training data pool can reveal privacy vulnerabilities, a challenge often addressed through membership inference attacks (MIAs). Traditional MIAs typically require access to model internals or rely on computationally intensive shadow models. This paper proposes an efficient, interpretable and principled Bayesian inference method for membership inference. By analyzing post-hoc metrics such as prediction error, confidence (entropy), perturbation magnitude, and dataset statistics from a trained ML model, our approach computes posterior probabilities of membership without requiring extensive model training. Experimental results on synthetic datasets demonstrate the method's effectiveness in distinguishing member from non-member datasets. Beyond membership inference, this method can also detect distribution shifts, offering a practical and interpretable alternative to existing approaches.
\end{abstract}

\section{Introduction}

Machine learning models, particularly deep neural networks, are vulnerable to privacy attacks such as membership inference attacks (MIAs), which determine whether a specific data point was included in a model's training set \cite{Shokri2017, Ye2022EnhancedMIA, Carlini2022}. These attacks exploit the tendency of models to exhibit distinct behaviors (e.g. higher confidence or lower loss) on training data compared to unseen data, potentially compromising the confidentiality of sensitive datasets, such as those containing medical or financial records. State-of-the-art MIAs typically rely on extensive knowledge of the target model. For example, shadow model-based approaches \cite{Shokri2017} train multiple models to mimic the target's behavior, while others, e.g. the likelihood ratio attack (LiRA) by Carlini et al. \cite{Carlini2022}, leverage model outputs or gradients. These methods often induce significant computational costs or require access to model internals, limiting their applicability in scenarios where only model outputs are available.

We propose a new MIA method that leverages Bayesian inference for post-hoc analysis of trained model and datasets. Once a ML model, e.g. a neural network, has been trained on member datasets, we pass the test data through the trained ML model, and extract resulting metrics such as accuracy, entropy, perturbation magnitude, and dataset statistics, and uses these metrics to compute posterior probabilities of membership. This approach doesn't require access to a 'training' set, although known knowledge about member and non-member datasets can improve its performance. This post-hoc method is computationally efficient, interpretable, requires minimum model query and fine-tuning, making it well-suited for real-world deployment scenarios where privacy assessments are conducted after model training.

\section{Related Work}

Membership inference has been extensively studied, with early research in the 1990s and 2000s laying the groundwork for understanding privacy risks in statistical and machine learning contexts. \textbf{Dalenius (1977)} \cite{Dalenius1977towards} established foundational concepts of disclosure risks in aggregated data, addressing vulnerabilities in statistical databases. \textbf{Agrawal and Srikant (2000)} \cite{Agrawal2000privacy} proposed one of the first privacy-preserving data mining frameworks, addressing individual data leakage, while \textbf{Homer et al. (2008)} \cite{Homer2008Resolving} demonstrated the feasibility of inferring individual genomic membership from statistical aggregates, raising awareness of privacy vulnerabilities that later extended to modern machine learning models. \textbf{Shokri et al. (2017)} \cite{Shokri2017} introduced membership inference attacks (MIAs), using multiple shadow models to replicate a target model's behavior and training a binary classifier for membership inference; this approach achieves high accuracy but is computationally intensive due to the need for multiple model trainings. \textbf{Ye et al. (2022)} \cite{Ye2022EnhancedMIA} enhanced this framework by incorporating features like loss distributions and confidence scores, improving attack success rates for deep neural networks in both black-box and white-box scenarios, though it still relies on shadow models and requires curated auxiliary datasets. \textbf{Carlini et al. (2022)} \cite{Carlini2022} developed the likelihood ratio attack (LiRA), employing a hypothesis-testing framework to compare membership likelihoods based on model outputs, excelling at low false-positive rates; however, LiRA demands multiple model queries and significant computational resources. \textbf{Hu et al. (2022)} \cite{Hu2022Survey} surveyed MIA techniques, categorizing them by strategies (e.g. shadow models, threshold-based, metric-based) and access levels, highlighting trade-offs between accuracy, computational cost, and model access requirements, and emphasizing the need for efficient and interpretable methods.

While these methods advance membership inference capabilities, their reliance on shadow models, extensive model queries, or model internals limits scalability and efficiency. Our Bayesian inference-based approach addresses these limitations by querying the trained model and computing post-hoc metrics, offering efficiency and interpretability.

\section{The Bayesian approach to membership inference}

We aim to infer whether a test dataset was part of the training set (member) or not (non-member) using Bayesian belief updating. To achieve this, we define a prior probability for the membership status, representing our initial belief about its membership status, compute likelihoods based on metrics extracted from the trained ML model, and update the posterior membership distribution. This approach is computationally efficient, as it avoids training an additional observer/assessor model, and provides an interpretable probabilistic framework for membership inference.

\paragraph{The problem and Bayesian solution}

In a supervised learning problem, we have a trained machine learning model $f(X, \theta)$, where $X$ are the inputs and $\theta$ represents the model parameters (e.g. weights in a neural network). We aim to determine whether a test dataset $D_{\text{test}} = (X_{\text{test}}, y_{\text{test}})$ was used in training the machine learning model, i.e. whether $D_{\text{test}}$ is a member of the training set, denoted as $M=1$, versus a non-member, $M = 0$.

Bayesian inference provides a framework for updating our belief about whether a dataset was part of the training set by combining a prior belief with likelihoods derived from observed model behavior metrics. For a binary hypothesis $M \in \{0, 1\}$, where $M$ indicates membership status, as per Bayes' theorem, the posterior probability given data $D$ is:
\begin{equation} \label{eq:posterior_membership}
    p(M = 1 | D) = \frac{p(D | M = 1) p(M = 1)}{p(D)}
\end{equation}
where $p(M = 1)$ is the prior probability of membership, $p(D | M = 1)$ is the likelihood of observing the data given the membership status $M=1$, and $p(D)$ is the marginal likelihood (i.e. normalizing constant, also termed \textit{evidence}), which is computed as $p(D) = p(D | M = 1) p(M = 1) + p(D | M = 0) p(M = 0)$.

We can plug the test dataset $D_{\text{test}} = (X_{\text{test}}, y_{\text{test}})$ into the trained model $f(X, \theta)$, compute the likelihoods, and obtain the posterior $p(M = 1 | D_{\text{test}})$. Computing the posterior, in many cases, can be intractable due to complex likelihoods or computational restrictions (e.g. large volume of or high-dimensional data); approximate methods such as Markov Chain Monte Carlo (MCMC \cite{Bayesian_signal_processing_Joseph}) or variational inference \cite{Jordan1999introduction} may be employed to estimate the posterior effectively.

If we already know some subset of the training data $D_{\text{train}} = (X_{\text{train}}, y_{\text{train}})$ has been used in training the machine learning model $f(X, \theta)$, we can utilize this information to inform our test on the unknown test set $D_{\text{test}} = (X_{\text{test}}, y_{\text{test}})$. Specifically, we can analyze the model's behavior on $D_{\text{train}}$ to calibrate the prior $p(M = 1)$ and the likelihood model $p(D | M)$, thereby improving the accuracy of the posterior probability estimation for $D_{\text{test}}$. For example, by extracting metrics such as prediction accuracy, entropy, and perturbation magnitude from $D_{\text{train}}$, we can empirically estimate the expected behavior of the model on member datasets, allowing us to set more informed parameters for the likelihood model (e.g. the mean and variance of Gaussian distributions). Additionally, this prior knowledge can help adjust the prior probability $p(M = 1)$ based on the proportion of known member datasets relative to the total dataset population and the similarity between the training and test sets, leading to a more robust and data-driven inference process. Alternatively, we can use information from both training and test sets, i.e. $D = D_{\text{train}} \cup D_{\text{test}}$, in the Bayesian inference as per Eq.\ref{eq:posterior_membership}. By combining these datasets, we can compute joint metrics that capture the relative behavior of the model across both sets, such as the difference in prediction accuracy or entropy between $D_{\text{train}}$ and $D_{\text{test}}$, which may reveal patterns indicative of membership. This joint approach can also facilitate the modeling of dependencies between the datasets, potentially using a hierarchical Bayesian model to account for shared characteristics, and improve the estimation of the marginal likelihood $p(D)$ by incorporating a broader data context, thus enhancing the overall accuracy and robustness of the membership inference process.

\paragraph{The membership inference procedure} 
For a test dataset, we first assign priors $p(M = 1)$ and $p(M = 0)$ to the membership status $M$, reflecting our initial belief about the membership status before observing the test dataset. A natural choice, given no prior knowledge, is a uniform (vague or unbiased) prior, modeling $M$ as a Bernoulli random variable with $p(M = 1) = p(M = 0) = 0.5$, assuming equal likelihood for the dataset being a member or non-member. Alternatively, if prior knowledge suggests a different balance, an moer informed prior, e.g. $p(M = 1) = 0.3$, can be used.

Then we plug the test dataset $D_{\text{test}} = (X_{\text{test}}, y_{\text{test}})$ into the trained model $f(X,\theta)$ and calculate following metrics:
\begin{itemize}
    \item \textit{Prediction accuracy}: the proportion of correct predictions, e.g. classification accuracy or MSE in regression. Classification for example, $\text{acc} = \frac{1}{n} \sum_{i=1}^n \mathbb{I}(\hat{y}_i = y_i)$, where $\hat{y}_i$ is the predicted label for the $i$-th sample. The associated \textit{prediction error metric} = $1- \text{acc}$.
    \item \textit{Prediction entropy}: the average entropy of predicted probabilities, implying \textit{confidence}. Classification for example, $H = \frac{1}{n} \sum_{i=1}^n \text{entropy}(p_i)$, where $p_i$ is the softmax probability distribution over classes for the $i$-th sample.
    \item \textit{Perturbation magnitude}: sensitivity of model weights to the dataset, computed as the $L_2$ norm of the difference between the original weights and weights after finetuning \footnote{We re-train the trained model using a specific dataset (e.g. the test set) with small number of epochs (e.g. 5 epochs).} the model on $D_{\text{test}}$, denoted as \textit{pert}.
    \item \textit{Dataset statistics}: e.g. mean $\mu_{\text{feat}}$ and variance $\sigma^2_{\text{feat}}$ of input features, computed across all features.
\end{itemize}
These metrics form a feature vector $\mathbf{z}=[1-\text{acc}, H, \text{pert}, \mu_{\text{feat}}, \sigma^2_{\text{feat}}]$, which encapsulates the behavior of the neural network on $D_{\text{test}}$.

The likelihood $p(D_{\text{test}} | M)$ is modeled through the metrics $\mathbf{z}$. We assume each metric, or a derived quantity, follows a standard normal distribution \footnote{We can also assign priors on the likelihood model parameters, which leads to a hierarchical Bayesian model.} conditioned on the membership status. The resulting likelihood is a product \footnote{This product may be factorised with further independence assumptions.} of Gaussian distributions. This likelihood model is based on our expectation that these metrics exhibit different behaviors for member versus non-member datasets. Specifically, for member datasets ($M=1$), the neural network is expected to have lower prediction errors, lower entropy (higher confidence), and smaller weight perturbations, as the model was trained on similar data; for non-member datasets ($M = 0$), higher errors, higher entropy, and larger perturbations are expected due to distributional differences.

Using 3 metrics for example, denoting $z_1 = 1 - \text{acc}$ (prediction error), $z_2 = H$ (entropy), and $z_3 = \text{pert}$ (perturbation magnitude), we assume each metric variable follows a Gaussian distribution conditioned on $M$:
\[
z_i | M \sim \mathcal{N}(\mu_{i,M}, \sigma^2_{i,M}), \quad i = 1, 2, 3
\]
where $\mu_{i,M}$ and $\sigma^2_{i,M}$ are the mean and variance of the $i$-th metric given the membership status $M$. We assume these metrics are Gaussian distributed, with means $\mu_{i,M}$ and variances $\sigma^2_{i,M}$ being calibrated using e.g. known 'training' sets \footnote{In practice and later experiments, the means and variances are derived using known 'training' datasets (both member and non-member datasets) to reflect expected differences (lower for members, higher for nonmembers).}. For example, for $M=1$, we can set $\mu_{1,1} = 0$, $\mu_{2,1} = 0$, $\mu_{3,1} = 0$ (lower errors, entropy, and perturbations); for $M = 0$, we can set $\mu_{1,0} = 1$, $\mu_{2,0} = 1$, $\mu_{3,0} = 1$ (higher errors, entropy, and perturbations, shifted by 1 standard deviation). Then, the likelihood for each metric is:
\[
p(z_i | M) = \frac{1}{\sqrt{2\pi}} \exp\left(\frac{(z_i - \mu_{i,M})^2}{2}\right)
\]
Further assuming independence among the metrics (a simplified assumption for tractability \footnote{This independence assumption among metrics may oversimplify the problem, as metrics like accuracy and entropy may be correlated.}), the joint likelihood becomes:
\[
p(\mathbf{z} | M) = \prod_{i=1}^3 p(z_i | M)
\]

Finally, we apply Bayes' theorem to compute the posterior probability (Eq.\ref{eq:posterior_membership}) of membership: 
\[
p(M = 1 | \mathbf{z}) = \frac{p(\mathbf{z} | M = 1) p(M = 1)}{p(\mathbf{z})}
\]
with the marginal likelihood computed as $p(\mathbf{z}) = p(\mathbf{z} | M = 1) p(M = 1) + p(\mathbf{z} | M = 0) p(M = 0)$. Substituting the prior ($p(M = 1) = p(M = 0) = 0.5$) and the likelihoods, we calculate $p(M = 1 | \mathbf{z})$, which represents the updated belief about the membership status of $D_{\text{test}}$. This posterior probability serves as the membership inference score, with higher values indicating a higher likelihood of the dataset being a member.

In summary, for each test dataset, we first extract the metrics $\mathbf{z}$ using the trained ML model, i.e. the test data is passed to the model oracle and outputs are collected. Then we compute the likelihoods $p(\mathbf{z} | M = 1)$ and $p(\mathbf{z} | M = 0)$. Finally we apply Bayes' theorem to obtain the posterior $p(M = 1 | \mathbf{z})$. This procedure is formalized in Algorithm.\ref{algo:Bayesian_inference_of_membership}.

\begin{algorithm}[H]
\caption{Bayesian inference-based training dataset membership inference}
\label{algo:Bayesian_inference_of_membership}
\begin{algorithmic}[1]
\Require Trained neural network $f$, test dataset $D_{\text{test}} \in \mathbb{R}^{n \times d}$
\Ensure Posterior probability $p(M = 1 | D_{\text{test}})$
\State Pass $D_{\text{test}}$ through $f$ to obtain predictions
\State Compute metrics: prediction error $z_1$, entropy $z_2$, perturbation magnitude $z_3$, data statistics $z_4$, etc
\State Assign priors, e.g. $p(M = 1) = p(M = 0) = 0.5$
\State Compute likelihoods $p(\mathbf{z} | M = 1)$ and $p(\mathbf{z} | M = 0)$ using Gaussian models
\State Compute marginal likelihood $p(\mathbf{z}) = p(\mathbf{z} | M = 1) p(M = 1) + p(\mathbf{z} | M = 0) p(M = 0)$
\State Compute posterior $p(M = 1 | \mathbf{z}) = \frac{p(\mathbf{z} | M = 1) p(M = 1)}{p(\mathbf{z})}$
\State \Return $p(M = 1 | \mathbf{z})$
\end{algorithmic}
\end{algorithm}

\paragraph{Complexity} 
Computational burden is dominated by the fine-tuning step for computing perturbation magnitude, which involves $E$ epochs of training on $n$ samples with a model $f$ having $m$ parameters, resulting in a complexity of $O(E \cdot n \cdot m)$. Other steps, including forward passes ($O(n \cdot m)$), metric computations like accuracy and entropy ($O(n)$), dataset statistics ($O(n \cdot d)$), and likelihood/posterior calculations ($O(1)$), are less significant, resulting in the overall \textit{time complexity} $O(E \cdot n \cdot m)$, with \textit{space complexity} $O(m + d)$.

\section{A synthetic classification dataset example}

\subsection{Experimental Procedure}

We created \footnote{All synthetic data were generated using the \texttt{scikit-learn} \texttt{make\_classification} function \cite{sklearn2011}.} a synthetic dataset for a binary classification task. The member dataset comprises 10 subsets, each with 200 data points (2000 total), sampled from a balanced (50\% \textit{vs.} 50\%) two-cluster Gaussian mixture with a class separation of 1.0. Each data point has 10 features, all of which are informative. The non-member dataset includes 10 subsets (2000 data points total) with a larger class separation of 5.0, increased noise (flip probability of 0.2), and imbalanced classes (80\% \textit{vs.} 20\%). A three-layer \textit{multi-layer perceptron} (MLP) with 16 hidden units was trained on the combined member dataset (2000 data points) for 100 epochs using the \textit{Adam} optimizer with a learning rate of 0.01. After training, all datasets were processed through the MLP to extract 23 metrics: \textit{prediction error}, \textit{average entropy of predicted probabilities}, \textit{perturbation magnitude} (L2 norm of the difference between original and fine-tuned weights after 5 epochs of fine-tuning the trained model), and the \textit{means and variances} of each of the 10 features.

To calibrate the likelihood model, we computed the metrics for all member and non-member datasets. The means of these metrics for member datasets were used as $\mu_{i,1}$, and for non-member datasets as $\mu_{i,0}$, with $i$ ranging over the 23 metrics (prediction error, entropy, perturbation magnitude, and the 10 feature means and 10 feature variances). The standard deviations $\sigma_i$ were computed as the pooled standard deviation across both member and non-member datasets for each metric. Two experiments were conducted to evaluate the Bayesian inference method. In the \textbf{first} experiment, we applied Bayesian inference to 3 test sets, each containing 200 points: a \textbf{subset} of the member dataset, a \textbf{re-sampled} dataset with the same distribution as the member dataset (with class separation 1.0, same used in generating the member datasets), and a \textbf{distinct} dataset with intermediate characteristics (class separation 3.0, noise 0.2, imbalanced classes 80\% \textit{vs.} 20\%). For each test set, we computed the metrics, modeled their likelihoods as a product of 23 Gaussian distributions conditioned on membership status using the calibrated $\mu_{i,1}$, $\mu_{i,0}$, and $\sigma_i$ (calculated from the member and non member training sets \footnote{The 3 'training' sets, i.e. member datasets, non-member datasets, and similar non-member datasets, are \textbf{only} used to calibrate $\mu_{i,1}$, $\mu_{i,0}$, and $\sigma_i$ which are used in the likelihood model; if there are no known member or non-member datasets available, one can manually set these values, e.g. $\mu_{1,1} = 0$, $\mu_{2,1} = 0$, $\mu_{3,1} = 0$ (lower errors, entropy, and perturbations), and $\mu_{1,0} = 1$, $\mu_{2,0} = 1$, $\mu_{3,0} = 1$ (higher errors, entropy, and perturbations, shifted by 1 standard deviation).}), and calculated the posterior probability of membership using a uniform prior $p(M = 1) = p(M = 0) = 0.5$. In the \textbf{second} experiment, we further include the similar non-member datasets to calculate the empirical statistics $\mu_{i,1}$, $\mu_{i,0}$ and $\sigma_i$. We repeated the inference on the \textbf{same} 3 test sets to assess the consistency of the method. In both experiments, the member datasets that are used to train the ML model (the shallow neural net) remain the same. We examine the posterior membership probabilities of these 3 test sets across scenarios.

\subsection{Results}

\begin{figure}[H]
  \centering
  \begin{subfigure}[b]{0.40\textwidth}
    \centering
    \includegraphics[width=\linewidth]{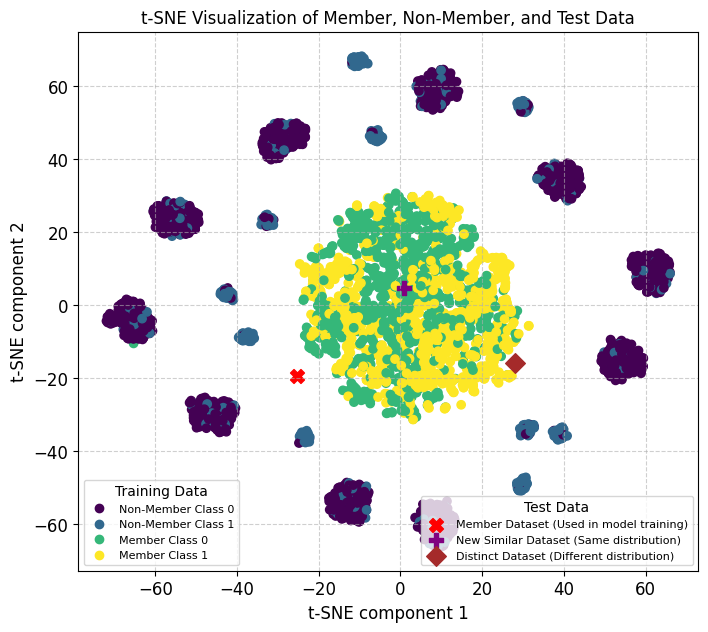}
    \caption{2D t-SNE visualisation of data}
    \label{fig:experiment1_tsne}
  \end{subfigure}
  \hspace{1cm}
  \begin{subfigure}[b]{0.4\textwidth}
    \centering
    \includegraphics[width=\linewidth]{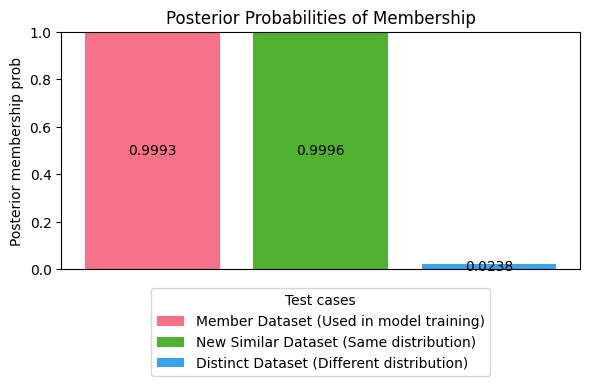}
    \caption{Inferred membership probabilities for 3 test datasets}
    \label{fig:experiment1_posteriors}
  \end{subfigure}
  \caption{Experiment 1 - synthetic data and results.}
  \label{fig:experiment1}
\end{figure}

\begin{figure}[H]
  \centering
  \begin{subfigure}[b]{0.40\textwidth}
    \centering
    \includegraphics[width=\linewidth]{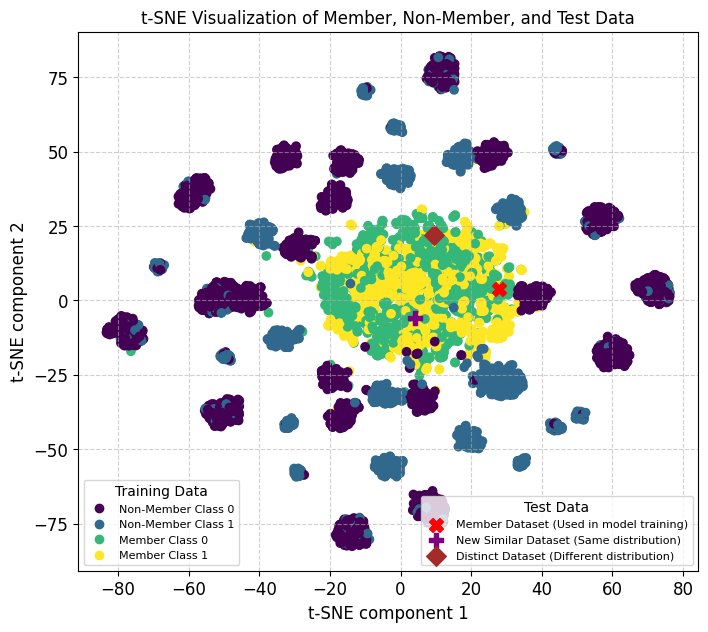}
    \caption{2D t-SNE visualisation of data}
    \label{fig:experiment2_tsne}
  \end{subfigure}
  \hspace{1cm}
  \begin{subfigure}[b]{0.4\textwidth}
    \centering
    \includegraphics[width=\linewidth]{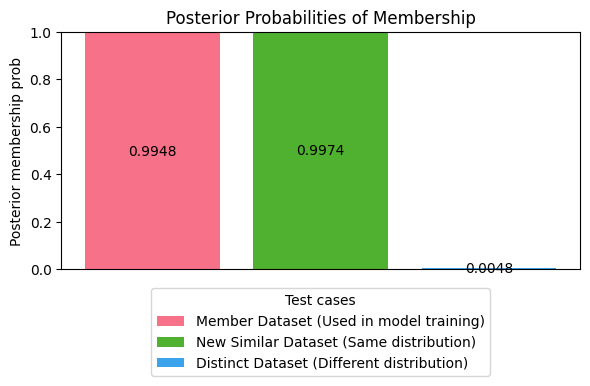}
    \caption{Inferred membership probabilities for 3 test datasets}
    \label{fig:experiment2_posteriors}
  \end{subfigure}
  \caption{Experiment 2 - synthetic data and results.}
  \label{fig:experiment2}
\end{figure}

We observe from the dimension reduced visualizations of the combined member and non-member datasets, along with the test sets in Fig.\ref{fig:experiment1_tsne} and Fig.\ref{fig:experiment2_tsne}, that member data formed a dense central cluster, while non-member data was more scattered. In the first experiment, the distinct test set was positioned slightly outside the member cluster, indicating partial overlap due to its intermediate separation distance. In the second experiment, the inclusion of similar non-member data brought the distinct test set closer to the member cluster, reflecting increased overlap due to shared characteristics with the similar non-member dataset. 

In the first experiment, we used 2 datasets to calculate the empirical statistics $\mu_{i,1}$, $\mu_{i,0}$, and $\sigma_i$ for the likelihood model: a member dataset and a distinct non-member dataset, as visualized in Fig.\ref{fig:experiment1_tsne}. The inferred posterior probabilities for three test datasets, i.e. a retrieved member dataset, a new similar dataset, and a distinct dataset, are shown in Fig.\ref{fig:experiment1_posteriors}. We observe that, the Bayesian inference method effectively distinguished the member-like datasets from the distinct dataset. The posterior probability of membership for the member dataset is 0.9993, and for the new similar dataset, it is 0.9996, reflecting the strong alignment of their metrics with the calibrated likelihood model for $M = 1$. The distinct test dataset, with metrics deviating due to its different data generating process, yielded a posterior probability of 0.0238, indicating a very low likelihood of membership. This evidenced separation demonstrates the method's effectiveness when non-member data significantly differs from member data (class separation 5.0 in the non-member set), enhanced by the empirical calibration of the likelihood parameters across all 23 metrics.

In the second experiment, we included an additional similar non-member dataset (class separation 3.0, balanced classes) alongside the member and distinct non-member datasets to calibrate the likelihood model parameters, as visualized in Fig.\ref{fig:experiment2_tsne}. The inferred posterior probabilities for the same three test datasets are shown in Fig.\ref{fig:experiment2_posteriors}. The inclusion of the similar non-member dataset slightly affected the calibrated metrics in the likelihood model and therefore the inference outcomes. The posterior probabilities for the member dataset and new similar dataset remained very high but decreased marginally to 0.9948 and 0.9974, respectively, reflecting a minor reduction in confidence due to the increased similarity between member and non-member distributions in the likelihood model. The distinct dataset's posterior probability decreased further to 0.0048, as its characteristics (class separation 3.0) now more closely matched the similar non-member data used in calibration, making the model more certain that datasets with these characteristics are non-members (informed by the derived metrics $\mu_{i,0}$ and $\sigma_i$ from both the distinct and similar, non-member datasets), thus enhancing the discriminative power for distinguishing truly distinct datasets from members. Both experiments demonstrate the method's consistent ability to differentiate membership status, with certain sensitivity to the inclusion of non-member datasets with distributions similar to the test data.

\subsection{Discussion}

The Bayesian inference method employs a comprehensive set of 23 metrics, including prediction error, entropy, perturbation magnitude, and the means and variances of all 10 features, to effectively capture differences in the model behaviors between member and non-member datasets. The use of a uniform prior ensures an unbiased starting point, while the Gaussian likelihood model, calibrated with empirical statistics from known 'training' data, provides a tractable way to update beliefs. Unlike shadow model-based methods, our approach requires only a trained ML model and small budget (e.g. 5 epochs) fine-tuning, making it computationally efficient with a time complexity of $O(E \cdot n \cdot m)$, where $E$ is the number of fine-tuning epochs, $n$ is the number of samples, and $m$ is the number of model parameters. The posterior probabilities offer a clear probabilistic interpretation, which is useful in practical deployment and decision-making regarding privacy risks.

The experiments demonstrate the method's robust discriminative power while revealing a nuanced sensitivity to the non-member dataset distribution. In the first experiment, the significant separation between member and non-member data (class separation 5.0) enabled near-certain differentiation, with member-like datasets achieving posterior probabilities exceeding 0.999, whereas the distinct test set's intermediate distance (class separation 3.0) resulted in a very low probability of 0.0238, indicating minimal overlap with the member distribution. In the second experiment, incorporating similar non-member data (class separation 3.0) slightly lowered the probabilities for member-like datasets to approximately 0.995, reflecting a minor increase in uncertainty due to overlapping distributions. Conversely, the distinct dataset's probability decreased further to 0.0048, showing the method's enhanced confidence in classifying such datasets as non-members, as the calibration now better captured the distribution of such datasets.

The advantages of this Bayesian method lie in: (1) It avoids training an additional observer or assessor model, reducing computational overhead while providing a probabilistic interpretation of membership. (2) It provides a clear probabilistic framework, with the posterior $p(M = 1 | \mathbf{z})$ directly interpretable as the probability of membership. (3) Prior beliefs can be incorporated and different likelihood models reflected in the Bayesian framework, making it flexible if more information about the data and trained model is available. (4) Flexibility can be extended by incorporating dependencies among metrics (e.g. a covariance structure) using, for example, multivariate Gaussian likelihoods. (5) Constant updating when new data becomes available, in a similar fashion to Bayesian online learning, allows the model to iteratively refine its posterior probabilities by incorporating new observations, enabling adaptability to evolving datasets while maintaining consistency with prior knowledge.

We also outline some limitations including: (1) the Gaussian model for likelihoods may not perfectly fit the true distribution of the metrics; alternative distributions, e.g. log-normal for perturbation magnitude, can be used. (2) The choice of prior and likelihood parameters ($\mu_{i,M}$, $\sigma_i$) can largely impact the results, careful empirical calibration is desired. (3) The independence assumption among metrics may oversimplify the problem, as metrics like prediction error and entropy, or feature means and variances, may be correlated; introducing a covariance structure among the metrics could improve accuracy. 

\section{Conclusions}

This study introduces a Bayesian inference-based methodology for membership inference, utilising post-hoc analysis of a trained machine learning model's behavior to distinguish member from non-member datasets. By querying the trained model, collecting outputs, extracting predictive metrics, applying Bayesian updating with prior beliefs and empirically calibrated Gaussian likelihoods, our approach achieves robust performance without requiring extensive model training. Experiments on synthetic datasets demonstrate the method's effectiveness: in both experiments, the proposed method accurately identifies member-like datasets with near-certain posterior probabilities while assigning a very low probability to distinct datasets, with consistent and improved ability to classify members and non-members when more experiences become available for calibration. Additionally, this work has the potential of being applied to detect distributional shifts, enabling the identification of changes in data distributions that may impact model performance or impose privacy risks.

\paragraph{Data and code availability} The synthetic data and codes used in this work are publically available at \href{https://github.com/YongchaoHuang/Bayesian_inference_of_training_dataset_membership}{this github repository}.

\bibliographystyle{plain}
\bibliography{references}

\begin{thebibliography}{10}

\bibitem{Agrawal2000privacy}
Rakesh Agrawal and Ramakrishnan Srikant.
\newblock Privacy-preserving data mining.
\newblock In {\em Proceedings of the 2000 ACM SIGMOD International Conference on Management of Data}, pages 439--450. ACM, 2000.

\bibitem{Carlini2022}
Nicholas Carlini, Steve Chien, Milad Nasr, Shuang Song, Andreas Terzis, and Florian Tramèr.
\newblock Membership inference attacks from first principles.
\newblock In {\em 2022 IEEE Symposium on Security and Privacy (SP)}, pages 1897--1914, 2022.

\bibitem{Dalenius1977towards}
Tore Dalenius.
\newblock Towards a methodology for statistical disclosure control.
\newblock {\em Statistisk Tidskrift}, 15(429-444), 1977.

\bibitem{Bayesian_signal_processing_Joseph}
Joseph J.K. O Ruanaidh; William~J. Fitzgerald.
\newblock {\em Numerical Bayesian Methods Applied to Signal Processing}, page 244.
\newblock Springer, 1st edition, 1996.

\bibitem{Homer2008Resolving}
Nils Homer, Szabolcs Szelinger, Margot Redman, David Duggan, Waibhav Tembe, Jill Muehling, John~V. Pearson, Dietrich~A. Stephan, Stanley~F. Nelson, and David~W. Craig.
\newblock Resolving individuals contributing trace amounts of dna to highly complex mixtures using high-density snp genotyping microarrays.
\newblock {\em PLOS Genetics}, 4:1--9, 08 2008.

\bibitem{Hu2022Survey}
Hongsheng Hu, Zoran Salcic, Lichao Sun, Gillian Dobbie, Philip~S. Yu, and Xuyun Zhang.
\newblock Membership inference attacks on machine learning: A survey.
\newblock {\em ACM Comput. Surv.}, 54(11s), September 2022.

\bibitem{Jordan1999introduction}
Michael~I. Jordan, Zoubin Ghahramani, Tommi~S. Jaakkola, and Lawrence~K. Saul.
\newblock An introduction to variational methods for graphical models.
\newblock {\em Machine Learning}, 37:183--233, 1999.

\bibitem{sklearn2011}
Fabian Pedregosa, Ga{{\"e}}l Varoquaux, Alexandre Gramfort, Vincent Michel, Bertrand Thirion, Olivier Grisel, Mathieu Blondel, Peter Prettenhofer, Ron Weiss, Vincent Dubourg, Jake Vanderplas, Alexandre Passos, David Cournapeau, Matthieu Brucher, Matthieu Perrot, and {{\'E}}douard Duchesnay.
\newblock Scikit-learn: Machine learning in python.
\newblock {\em Journal of Machine Learning Research}, 12(85):2825--2830, 2011.

\bibitem{Shokri2017}
Reza Shokri, Marco Stronati, Congzheng Song, and Vitaly Shmatikov.
\newblock {Membership Inference Attacks Against Machine Learning Models}.
\newblock In {\em 2017 IEEE Symposium on Security and Privacy (SP)}, pages 3--18, Los Alamitos, CA, USA, May 2017. IEEE Computer Society.

\bibitem{Ye2022EnhancedMIA}
Jiayuan Ye, Aadyaa Maddi, Sasi~Kumar Murakonda, Vincent Bindschaedler, and Reza Shokri.
\newblock Enhanced membership inference attacks against machine learning models.
\newblock In {\em Proceedings of the 2022 ACM SIGSAC Conference on Computer and Communications Security}, CCS '22, page 3093–3106, New York, NY, USA, 2022. Association for Computing Machinery.

\end{thebibliography}

\end{document}